\documentclass{article}

\usepackage{arxiv}

\usepackage[utf8]{inputenc} % allow utf-8 input
\usepackage[T1]{fontenc}    % use 8-bit T1 fonts
\usepackage{hyperref}       % hyperlinks
\usepackage{url}            % simple URL typesetting
\usepackage{booktabs}       % professional-quality tables
\usepackage{amsfonts}       % blackboard math symbols
\usepackage{amsmath}
\usepackage{nicefrac}       % compact symbols for 1/2, etc.
\usepackage{microtype}      % microtypography
\usepackage{graphicx}
\usepackage{float}
\usepackage[numbers]{natbib}
\usepackage{doi}
\usepackage{titlesec}

\titlespacing*{\section}{0pt}{8pt plus 2pt minus 2pt}{4pt}
\titlespacing*{\subsection}{0pt}{6pt plus 2pt minus 2pt}{3pt}
\titlespacing*{\subsubsection}{0pt}{4pt plus 1pt minus 1pt}{2pt}

\renewcommand{\headeright}{}
\renewcommand{\undertitle}{}
\title{SENTIMENT ANALYSIS OF INDONESIAN SPOTIFY REVIEWS USING MACHINE LEARNING AND BILSTM}

\author{%
  \textbf{Uliano Wilyam Purba}\\
  Department of Data Science\\
  Faculty of Science\\
  Sumatera Institute of Technology\\
  Lampung, Indonesia\\
  \texttt{uliano.122450098@student.itera.ac.id}
  \And
  \textbf{Andre Hadiman Rotua Parhusip}\\
  Department of Data Science\\
  Faculty of Science\\
  Sumatera Institute of Technology\\
  \texttt{andre.122450108@student.itera.ac.id}
  \And
  \textbf{Sahid Maulana}\\
  Department of Data Science\\
  Faculty of Science\\
  Sumatera Institute of Technology\\
  \texttt{sahid.122450109@student.itera.ac.id}
  \And
  \textbf{Luluk Muthoharoh M. Si}\\
  Department of Data Science\\
  Faculty of Science\\
  Sumatera Institute of Technology\\
  \texttt{luluk.muthoharoh@sd.itera.ac.id}
  \And
  \textbf{Ardika Satria M. Si}\\
  Department of Data Science\\
  Faculty of Science\\
  Sumatera Institute of Technology\\
  \texttt{ardika.satria@sd.itera.ac.id}
  \And
  \textbf{Martin C.T. Manullang Ph.D.}\\
  Department of Informatics Engineering\\
  Faculty of Industrial Technology\\
  Sumatra Institute of Technology\\
  \texttt{martin.manullang@if.itera.ac.id}\\
}

\begin{document}
\maketitle
\small

\begin{abstract}
This paper compares three Scikit-learn classifiers --- Support Vector Machine (SVM), Multinomial Naive Bayes (MNB), and Decision Tree (DT) --- with a two-layer Bidirectional Long Short-Term Memory (BiLSTM) model for three-class sentiment classification of Indonesian Spotify reviews. From 100,000 scraped reviews, 70,155 cleaned samples are used for the machine learning track, while the BiLSTM is trained on a stratified 20,000-sample subset because training is performed in a CPU-only environment. Both tracks use the same preprocessing pipeline: slang normalization, stopword removal, and Sastrawi stemming. In 5-fold cross-validation, Decision Tree is the strongest classical model with weighted F1-score 0.7269, ahead of SVM (0.6958) and MNB (0.4619). On its held-out test set, BiLSTM reaches weighted F1-score 0.8069, but fails on the minority Neutral class (F1 = 0.000). These results show that BiLSTM provides stronger overall performance, whereas ML with SMOTE yields more balanced three-class recognition.
\end{abstract}


% ==============================================================================
\section{Introduction}
\label{sec:intro}

Sentiment analysis is valuable for app developers because user reviews reveal satisfaction, complaints, and feature requests at scale. For Indonesian reviews, this task is harder because the text is highly informal, rich in slang, and often inconsistent in spelling.

Spotify reviews from the Google Play Store provide a relevant case study for comparing classical machine learning and deep learning on Indonesian text. In this work, SVM, Multinomial Naive Bayes, and Decision Tree with TF-IDF are benchmarked against a two-layer BiLSTM. The comparison is practical rather than perfectly symmetric: classical models use the full 70,155 cleaned samples, while BiLSTM is trained on a 20,000-sample stratified subset due to CPU-only constraints.

This study contributes: (1) a shared Indonesian preprocessing pipeline for both tracks; (2) a 5-fold benchmark of three classical classifiers with TF-IDF and SMOTE; (3) evaluation of a compact BiLSTM with 1,096,579 parameters; and (4) a transparent comparison of overall and per-class performance, supported by public deployment of both models.

\begin{itemize}
    \item ML demo: \url{https://huggingface.co/spaces/tubespba-kelompoktuwir/sentimen-spotify-ml}
    \item DL demo: \url{https://huggingface.co/spaces/tubespba-kelompoktuwir/sentimen-spotify-dl}
\end{itemize}

% ==============================================================================
\section{Related Work}
\label{sec:related}

\subsection{Sentiment Analysis: Foundational Work}
Pang and Lee \cite{pang2008opinion} established the main tasks and evaluation principles of sentiment analysis. Their survey also showed that classical supervised models with bag-of-words or TF-IDF features remain strong baselines for document-level polarity classification.

\subsection{Classical Machine Learning for Text Classification}
TF-IDF combined with SVM, Multinomial Naive Bayes, and tree-based models is widely used in sentiment classification. Prior studies on app reviews and social media show that SVM is often strong, while performance still depends heavily on feature design, label balance, and language characteristics \cite{pratap2020comparative,rustam2021comparative}.

\subsection{Deep Learning for Sentiment Analysis}
LSTM and BiLSTM models are designed to capture sequential context more effectively than bag-of-words features. Prior work reports that BiLSTM performs well for sentiment tasks and can outperform classical methods on Indonesian text, although the gain usually comes with higher computational cost \cite{hochreiter1997lstm,xu2019bilstm,lin2023hybrid}.

\subsection{Indonesian NLP and Sentiment Analysis}
Indonesian NLP research has improved through resources such as IndoNLU and IndoBERT \cite{wilie2020indonlu}, yet sentiment analysis still faces challenges from informal spelling, slang, and morphology. Previous Indonesian studies also report that BiLSTM is competitive, especially when preprocessing is carefully tailored to the language \cite{kuncahyo2020bilstm,farizki2025bilstm,abiola2024sentiment,adriani2007stemming}.

\subsection{App Review Sentiment Analysis}
App-review sentiment analysis is an applied setting where deep models often perform well, but class imbalance remains a persistent issue \cite{ossai2024sentiment,chawla2002smote}. To the best of our knowledge, a direct benchmark between Scikit-learn classifiers and a PyTorch BiLSTM on large-scale Indonesian Spotify reviews with shared preprocessing has not been previously reported.

% ==============================================================================
\section{Dataset}
\label{sec:dataset}

\subsection{Data Source and Collection}
The dataset consists of 100,000 Indonesian-language Spotify reviews scraped from the Google Play Store using the {google-play-scraper} library. For modeling, only review text and star rating are used.

The raw dataset includes reviewer name, review text, rating, date, likes, and application version. Only the \textit{Ulasan} and \textit{Rating} columns are retained for sentiment classification.

\subsection{Sentiment Label Construction}
Following common rating-based labeling, ratings 1--2 are mapped to Negatif, rating 3 to Netral, and ratings 4--5 to Positif. This scheme matches the natural interpretation of app-store reviews.

\subsection{Dataset Statistics and Class Distribution}
\begin{figure}[!htbp]
\centering
\includegraphics[width=0.58\linewidth]{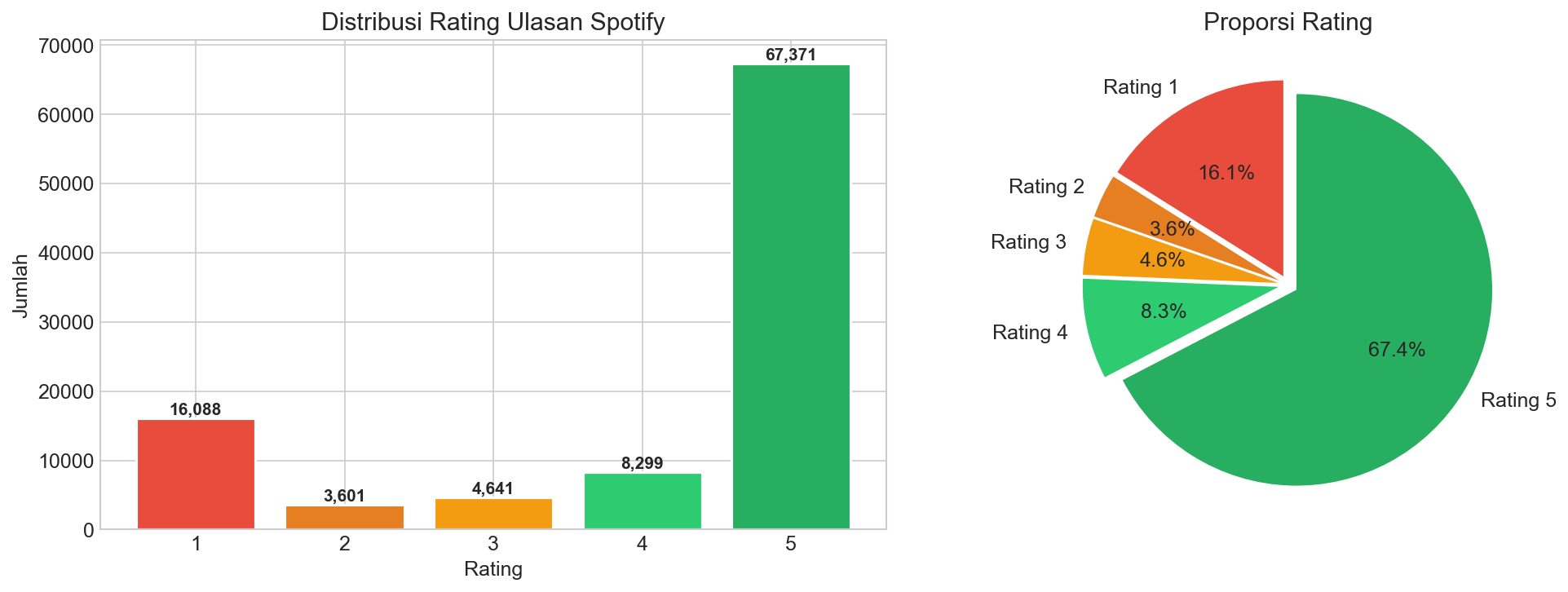} 
\caption{Sentiment distribution rating label from raw data.}
\label{fig:dist_raw}
\end{figure}

The dataset is strongly imbalanced, with positive reviews dominating the corpus. After label mapping, deduplication, and removal of empty reviews, the cleaned dataset contains 70,155 samples: Positif 47,991 (68.4\%), Negatif 18,026 (25.7\%), and Netral 4,138 (5.9\%).

\begin{table}[H]
\centering
\caption{Dataset Statistics and Class Distribution}
\label{tab:dataset_stats}
\begin{tabular}{lcccc}
\toprule
\textbf{Split} & \textbf{Positive} & \textbf{Netral} & \textbf{Negative} & \textbf{Total} \\ \midrule
Full cleaned dataset & 47,991 (68.4\%) & 4,138 (5.9\%) & 18,026 (25.7\%) & 70,155 \\
ML training set & 47,991 & 4,138 & 18,026 & 70,155 \\
DL subset (20K stratified) & 13,681 & 1,180 & 5,139 & 20,000 \\ \midrule
DL Train & 9,852 & 851 & 3,700* & 14,219* \\
DL Validation & $\sim$1,739 & $\sim$150 & $\sim$653* & 2,510* \\
DL Test & $\sim$2,090 & $\sim$179 & $\sim$786* & 2,953 \\ \bottomrule
\end{tabular}
\end{table}

Frequent negative themes include advertisements, premium access issues, and playback or lyric synchronization problems. These topics also appear in the cleaned vocabulary used by both tracks.

\subsection{Preprocessing Pipeline}
A unified preprocessing pipeline is applied to both tracks:
\begin{enumerate}
    \item {Lowercasing:} all text is converted to lowercase.
    \item {Noise removal:} URLs, mentions, hashtags, punctuation, symbols, and digits are removed.
    \item {Slang normalization:} 124 informal forms are mapped to standard Indonesian (e.g., gak $\rightarrow$ tidak).
    \item {Stopword removal:} 809 general and domain-specific stopwords are filtered out.
    \item {Morphological stemming:} tokens are reduced to root forms using Sastrawi [11].
\end{enumerate}

After preprocessing, the median review length is 4 tokens and the 95th percentile is 17 tokens. Therefore, the BiLSTM maximum sequence length is set to 17, and reviews that become empty after cleaning are removed.

\begin{table}[H]
\centering
\caption{Preprocessing Pipeline Summary}
\label{tab:preprocessing}
\begin{tabular}{lll}
\toprule
\textbf{Step} & \textbf{Method} & \textbf{Detail} \\ \midrule
Lowercasing & Python \texttt{str.lower()} & All characters converted to lowercase \\
Noise removal & Regex substitution & URLs, @mentions, punctuation, digits removed \\
Slang normalization & Custom dictionary & 124 entries (e.g., \textit{gak} $\rightarrow$ \textit{tidak}) \\
Stopword removal & Sastrawi + custom & 809 stopwords including domain terms \\
Morphological stemming & Sastrawi (ECS) & 40+ suffix/prefix rules for Indonesian \\ \bottomrule
\end{tabular}
\end{table}

% ==============================================================================
\section{Methodology}
\label{sec:method}

This study follows two parallel tracks --- machine learning (ML) and deep learning (DL) --- with the same preprocessing pipeline.

\subsection{Machine Learning Track (Scikit-learn)}
\subsubsection{Feature Extraction: TF-IDF}
Preprocessed reviews are converted into TF-IDF vectors with a maximum of 3,000 features \cite{pang2008opinion}.

\subsubsection{Class Balancing: SMOTE}
Because the dataset is imbalanced, SMOTE \cite{chawla2002smote} is applied only to the training fold in each cross-validation split.

\subsubsection{Classifiers}
Three classifiers are compared: SVM, Multinomial Naive Bayes, and Decision Tree. All are implemented in Scikit-learn and evaluated with stratified 5-fold cross-validation on the cleaned dataset.

\subsection{Deep Learning Track (BiLSTM)}
\subsubsection{Tokenization and Vocabulary}
Texts in the DL track are tokenized at word level. The training split produces a 3,285-token vocabulary, and all sequences are padded or truncated to length 17.

\subsubsection{Model Architecture}
The BiLSTM model uses 128-dimensional embeddings, a two-layer bidirectional LSTM with 128 hidden units per direction, dropout, and a fully connected classification head. The total parameter count is 1,096,579.

\subsubsection{Training Configuration}
The model is trained on a stratified 20,000-sample subset with Adam optimizer, learning rate $10^{-3}$, batch size 64, and unweighted cross-entropy loss. Early stopping selects the best checkpoint at epoch 3.

\begin{figure}[!htbp]
\centering
\includegraphics[width=0.62\linewidth]{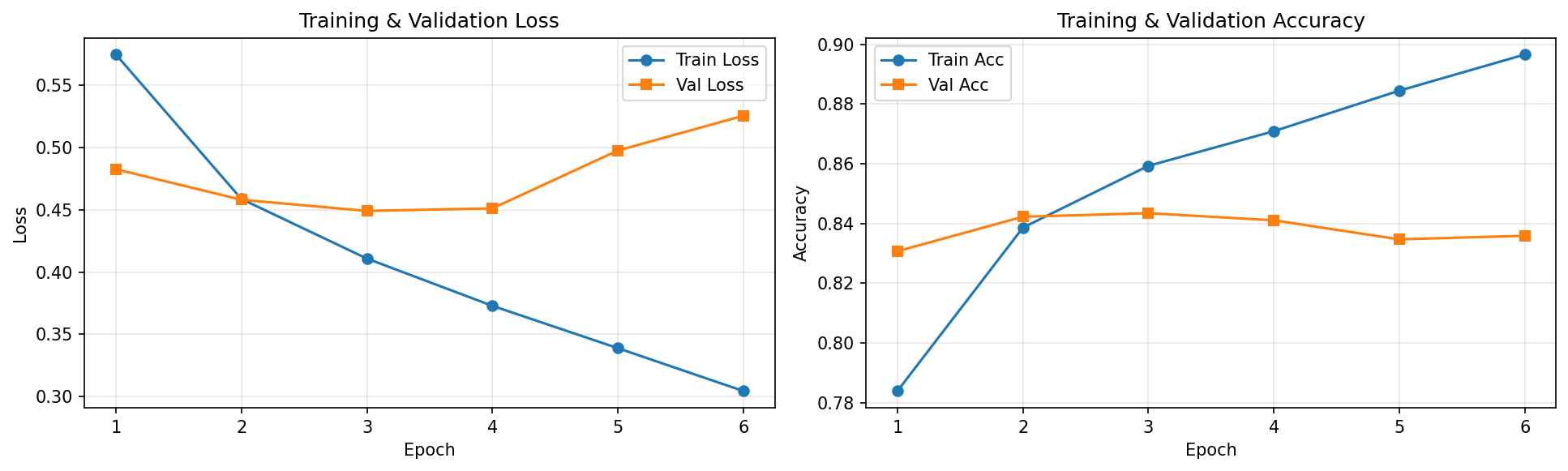}
\caption{Training and Validation Loss/Accuracy Curves during the optimization phase.}
\label{fig:curves_init}
\end{figure}

Using a 20,000-sample subset for the BiLSTM reflects CPU-only training constraints and is treated as a limitation in the comparison.

% ==============================================================================
\section{Experiments}
\label{sec:experiments}

\subsection{Experimental Setup}
All experiments are conducted in a CPU-only environment (Intel-based local machine, PyTorch 2.2.2+cpu). The ML track runs on the full cleaned dataset of 70,155 samples using Scikit-learn 1.4.2 with PyCaret as the AutoML orchestration framework. The DL track trains the BiLSTM on a stratified 20,000-sample subset using PyTorch 2.2.2. All random states are fixed at seed=42 across both tracks for reproducibility.

\subsection{ML Hyperparameter Configuration}

Table \ref{tab:ml_hyper} details the hyperparameter configuration for the three classical machine learning classifiers. All classifiers are evaluated via stratified 5-fold cross-validation on the full 70,155 training samples.

\begin{table}[H]
\centering
\caption{ML Classifier Hyperparameter Configuration}
\label{tab:ml_hyper}
\begin{tabular}{@{}lll@{}}
\toprule
\textbf{Classifier} & \textbf{Key Hyperparameters} & \textbf{Feature Representation} \\ \midrule
Support Vector Machine & Kernel: linear; C: 1.0 (default) & TF-IDF, max features=3,000 \\
Multinomial Naive Bayes & Alpha: 1.0 (Laplace smoothing) & TF-IDF, max features=3,000 \\
Decision Tree & Criterion: gini; splitter: best & TF-IDF, max features=3,000 \\ \bottomrule
\end{tabular}
\end{table}

All classifiers are evaluated via stratified 5-fold cross-validation on the full 70,155 training samples. Performance metrics are macro-averaged across folds. Following benchmark evaluation, the best-performing model (Decision Tree) undergoes hyperparameter tuning via PyCaret's tune\_model() with n\_iter=50, optimizing for macro F1-Score, before finalization with finalize\_model() (retraining on the full dataset).

\subsection{DL Hyperparameter and Training Configuration}
The BiLSTM is trained on a stratified 20,000-sample subset, split into training (14,219 / 72.1\%), validation (2,510 / 12.7\%), and test (2,953 / 15.0\%) sets using stratified partitioning to preserve class proportions at each split. The vocabulary is built exclusively from the training split (minimum token frequency = 2), yielding a final vocabulary of 3,285 unique tokens.

Table \ref{tab:dl_hyper} summarizes the training and architecture hyperparameters for the BiLSTM model.

\begin{table}[H]
\centering
\caption{BiLSTM Training Hyperparameters}
\label{tab:dl_hyper}
\begin{tabular}{@{}ll@{}}
\toprule
\textbf{Hyperparameter} & \textbf{Value} \\ \midrule
Optimizer & Adam \\
Learning rate ($\eta$) & $1\times10^{-3}$ \\
Loss function & CrossEntropy Loss (unweighted) \\
Batch size & 64 \\
Max epochs & 10 \\
Early stopping patience & 3 epochs (monitor: {val\_loss}) \\
Max sequence length & 17 tokens (95th percentile of training lengths) \\
Embedding dimension & 128 \\
LSTM hidden size & 128 per direction (256 total) \\
LSTM layers & 2 (bidirectional) \\
Dropout (post-LSTM) & 0.5 \\ \bottomrule
\end{tabular}
\end{table}

Training terminates via early stopping at epoch 6 (best checkpoint restored from epoch 3, val\_loss = 0.4491, val\_acc = 0.8434). The training dynamics are visualized in Figure 3 training loss decreases monotonically from 0.5747 to 0.3046 across six epochs, while validation loss reaches its minimum at epoch 3 (0.4491) before diverging — a classic overfitting signature on a small training subset. Validation accuracy plateaus between 0.83–0.84 from epoch 2 onward, indicating the model saturates its representational capacity on 14,219 training samples.

\subsection{Evaluation Metrics}
Both tracks are evaluated using accuracy, weighted precision, weighted recall, weighted F1-score, Cohen's Kappa, and macro F1. Weighted F1 is the primary metric because the dataset is imbalanced, while macro F1 is used to inspect minority-class performance. The ML track additionally reports AUC from 5-fold cross-validation, and the DL track reports the same metrics on the held-out test set.

\subsection{Fairness Considerations and Confounding Factors}
The two tracks are not fully symmetric: ML models train on 70,155 samples, while the BiLSTM uses 14,219 training samples. This reflects a practical CPU-only setting, but all preprocessing steps remain identical across both tracks.

% ==============================================================================
\section{Results and Discussion}
\label{sec:results}

\subsection{Machine Learning Results}

Table \ref{tab:ml_results} summarizes the 5-fold cross-validated performance of the three Scikit-learn classifiers on the full 70,155-sample corpus. Decision Tree achieves the highest scores across all primary metrics, attaining an accuracy of 0.7286, weighted F1-score of 0.7269, and Cohen's Kappa of 0.5928 --- indicating substantial agreement beyond chance. SVM ranks second with accuracy 0.6991 and weighted F1-score 0.6958. Multinomial Naive Bayes performs markedly lower, achieving accuracy 0.4948 and weighted F1-score 0.4619. The complete benchmark is presented below.

\begin{table}[H]
\centering
\caption{ML Cross-Validation Benchmark Results}
\label{tab:ml_results}
\begin{tabular}{lcccccc}
\toprule
\textbf{Model} & \textbf{Accuracy} & \textbf{AUC} & \textbf{Recall} & \textbf{Precision} & \textbf{F1-Score} & \textbf{Kappa} \\ \midrule
Decision Tree & 0.7286 & 0.8055 & 0.7286 & 0.7273 & 0.7269 & 0.5928 \\
Support Vector Machine & 0.6991 & --- & 0.6991 & 0.6956 & 0.6958 & 0.5487 \\
Multinomial Naive Bayes & 0.4948 & 0.6899 & 0.4948 & 0.6181 & 0.4619 & 0.2422 \\ \bottomrule
\end{tabular}
\end{table}

All metrics are macro-weighted averages from 5-fold stratified cross-validation on 70,155 samples with SMOTE applied within each fold. AUC not available (---) for SVM under the PyCaret configuration used.

Decision Tree performs better than SVM on this dataset, likely because SMOTE and short review length favor simpler partitioning over sparse linear separation. Multinomial Naive Bayes performs worst, indicating that feature independence is too restrictive for colloquial Indonesian reviews.

\begin{figure}[!htbp]
\centering
\includegraphics[width=0.62\linewidth]{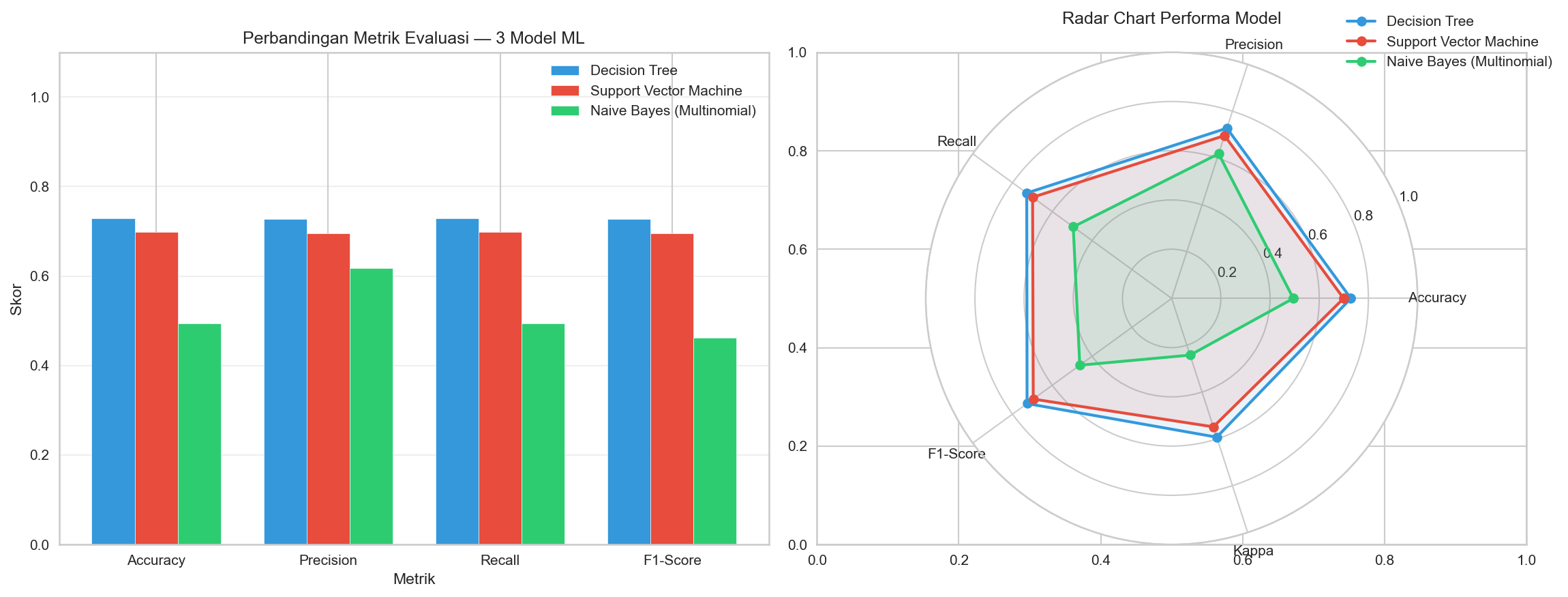}
\caption{Performance comparison of the three Scikit-learn classifiers across four evaluation metrics.}
\label{fig:ml_perf}
\end{figure}

The Decision Tree per-class report shows the best performance on Positif (F1 0.802), followed by Netral (0.717) and Negatif (0.707). This suggests that distinguishing Negative from Neutral remains the hardest case.

\begin{figure}[!htbp]
\centering
\includegraphics[width=0.3\linewidth]{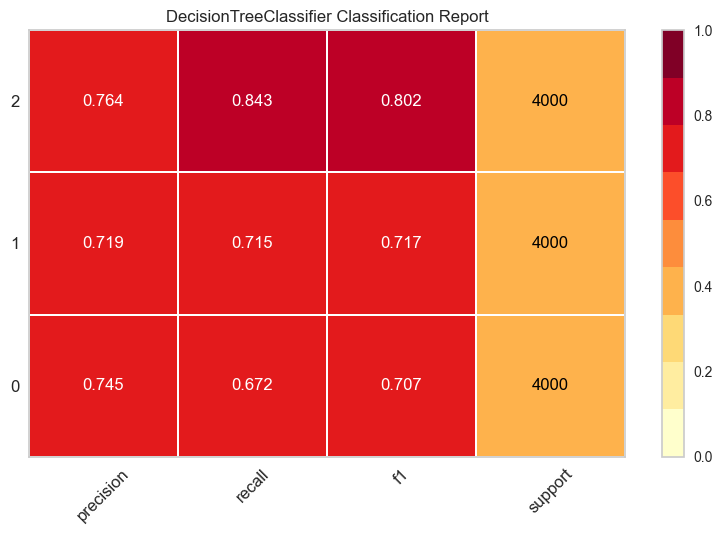}
\caption{Per-class classification report of the Decision Tree classifier on the held-out test fold.}
\label{fig:dt_report}
\end{figure}

The confusion matrix confirms that most errors occur between Negative and Neutral, which are adjacent in the rating-based labeling scheme.

\begin{figure}[!htbp]
\centering
\includegraphics[width=0.3\linewidth]{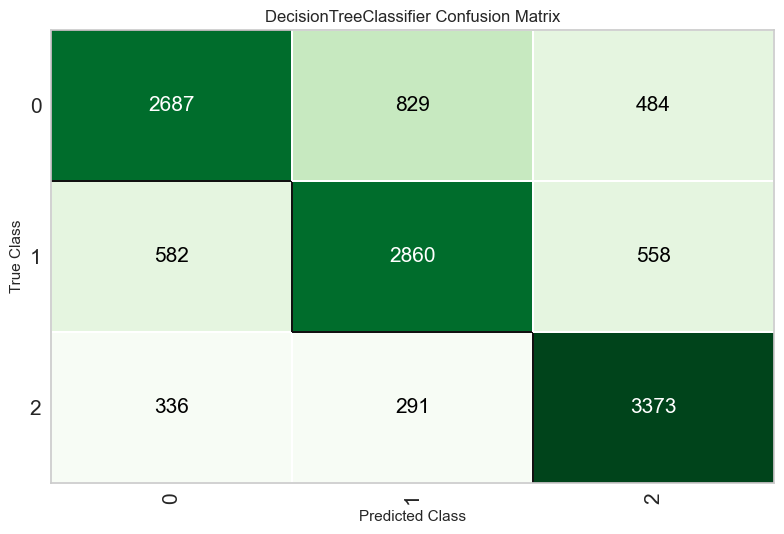}
\caption{Confusion matrix of the Decision Tree classifier on the test fold (balanced via SMOTE).}
\label{fig:dt_cm}
\end{figure}

\subsection{BiLSTM Deep Learning Results}

The BiLSTM model is evaluated on a held-out test set of 2,953 samples from the 20,000-sample subset. Table \ref{tab:bilstm_results} reports the metrics at the best checkpoint (epoch 3).

\begin{table}[H]
\centering
\caption{BiLSTM Test Set Evaluation Results}
\label{tab:bilstm_results}
\begin{tabular}{lc}
\toprule
\textbf{Metric} & \textbf{Value} \\ \midrule
Test Loss & 0.4735 \\
Accuracy & 0.8314 \\
Weighted Precision & 0.7841 \\
Weighted Recall & 0.8314 \\
Weighted F1-Score & 0.8069 \\
Macro F1-Score & 0.5478 \\
Total Parameters & 1,096,579 \\
Training Subset & 14,219 samples \\ \bottomrule
\end{tabular}
\end{table}

BiLSTM achieves weighted F1-score 0.8069, but its macro F1-score drops to 0.5478, indicating weak minority-class performance.

\begin{table}[H]
\centering
\caption{BiLSTM Per-Class Classification Report}
\label{tab:bilstm_report}
\begin{tabular}{lcccc}
\toprule
\textbf{Class} & \textbf{Precision} & \textbf{Recall} & \textbf{F1-Score} & \textbf{Support} \\ \midrule
Negatif & 0.7044 & 0.7807 & 0.7406 & 766 \\
Netral & 0.0000 & 0.0000 & 0.0000 & 176 \\
Positif & 0.8830 & 0.9234 & 0.9028 & 2,011 \\ \midrule
\textbf{Weighted Avg} & \textbf{0.7841} & \textbf{0.8314} & \textbf{0.8069} & \textbf{2,953} \\
\textbf{Macro Avg} & \textbf{0.5291} & \textbf{0.5680} & \textbf{0.5478} & \textbf{2,953} \\ \bottomrule
\end{tabular}
\end{table}

The per-class report shows a clear imbalance: BiLSTM performs very well on Positif (F1 0.9028), reasonably on Negatif (F1 0.7406), and fails completely on Netral (F1 0.0000). The confusion matrix confirms that none of the 176 Neutral samples are classified correctly.

\begin{figure}[!htbp]
\centering
\includegraphics[width=0.3\linewidth]{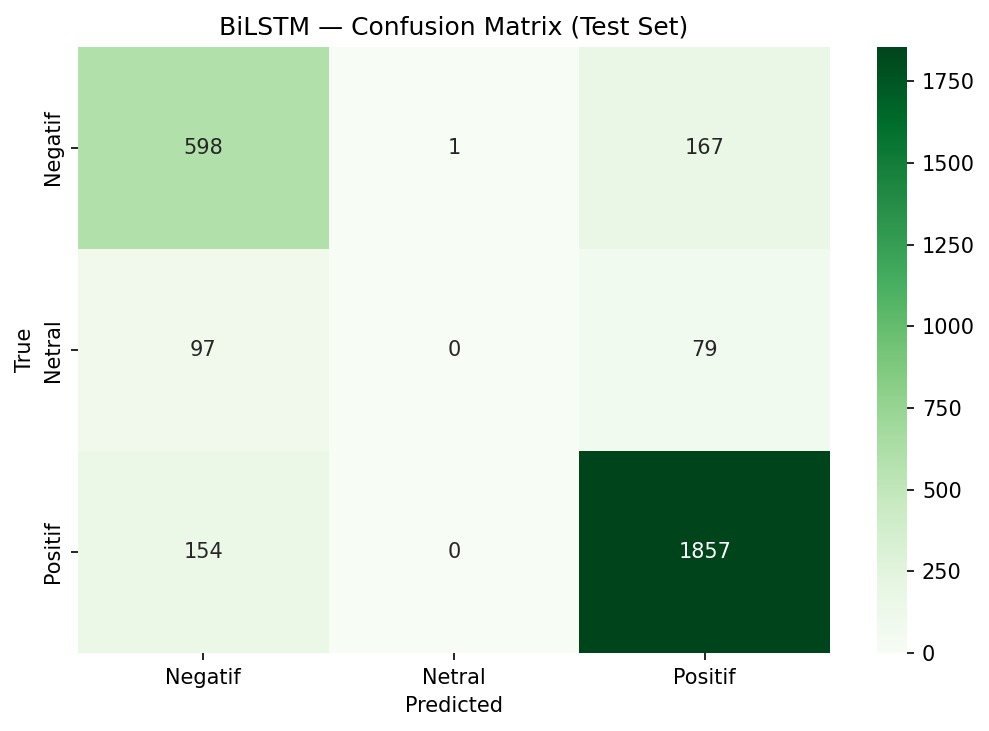}
\caption{Confusion matrix of the BiLSTM model on the held-out test set ($n = 2,953$).}
\label{fig:bilstm_cm}
\end{figure}

This Neutral-class collapse is mainly caused by class imbalance and the absence of class weighting or oversampling in the DL pipeline. The Neutral class is also semantically ambiguous, making it harder to separate from Negative and Positive in short informal reviews.

\begin{figure}[!htbp]
\centering
\includegraphics[width=1.0\linewidth]{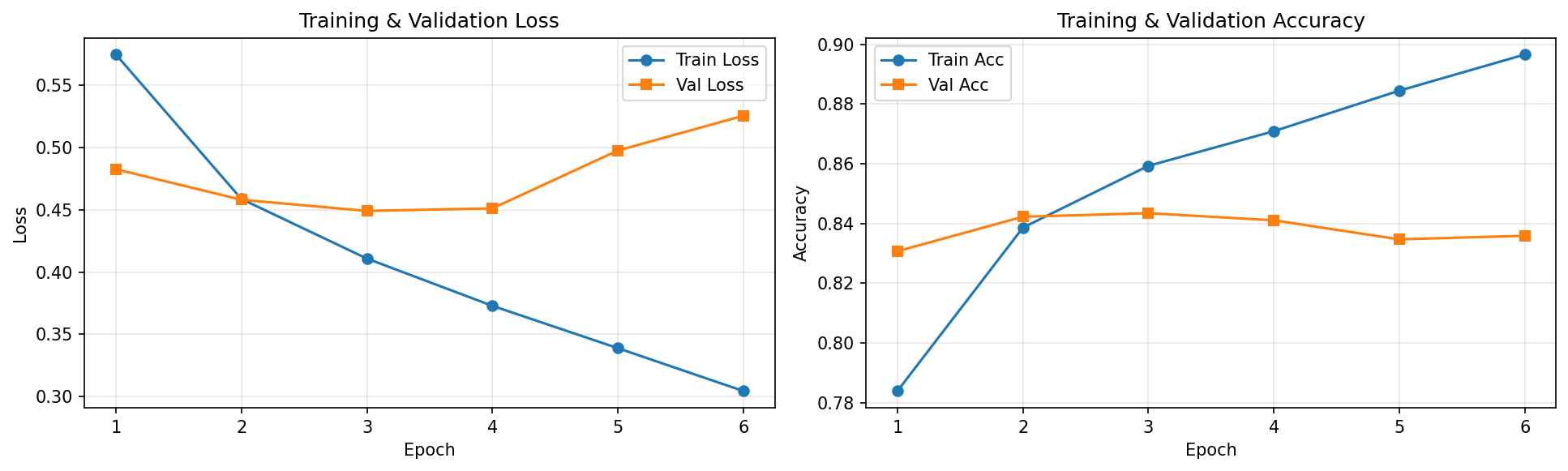}
\caption{BiLSTM training and validation loss (left) and accuracy (right) across six epochs.}
\label{fig:bilstm_curve}
\end{figure}

Figure \ref{fig:bilstm_curve} indicates early overfitting: training loss keeps decreasing, while validation loss reaches its minimum at epoch 3 and then rises. Validation accuracy stabilizes around 0.83--0.84, so early stopping selects epoch 3.

\subsection{Comparative Analysis: ML vs. BiLSTM}

\begin{table}[H]
\centering
\caption{Head-to-head performance comparison of ML and BiLSTM models.}
\label{tab:comparison}
\begin{tabular}{llcccccccc}
\toprule
\textbf{Model} & \textbf{Paradigm} & \textbf{Train Samples} & \textbf{Acc.} & \textbf{W-Prec.} & \textbf{W-Recall} & \textbf{W-F1} & \textbf{Macro-F1} & \textbf{Kappa} & \textbf{Params} \\ \midrule
Decision Tree & ML (sklearn) & 70,155 & 0.7286 & 0.7273 & 0.7286 & 0.7269 & $\sim$0.72* & 0.5928 & --- \\
SVM & ML (sklearn) & 70,155 & 0.6991 & 0.6956 & 0.6991 & 0.6958 & $\sim$0.70* & 0.5487 & --- \\
Mult. Naive Bayes & ML (sklearn) & 70,155 & 0.4948 & 0.6181 & 0.4948 & 0.4619 & $\sim$0.46* & 0.2422 & --- \\
BiLSTM & DL (PyTorch) & 14,219 & 0.8314 & 0.7841 & 0.8314 & 0.8069 & 0.5478 & --- & 1,096,579 \\ \bottomrule
\end{tabular}
\end{table}

Using weighted metrics, BiLSTM outperforms the best ML model. Its weighted F1-score is 0.8069 versus 0.7269 for Decision Tree, and its accuracy is 0.8314 versus 0.7286.

However, this advantage must be interpreted carefully because the two tracks are not symmetric. ML models are trained on 70,155 samples, whereas BiLSTM uses only 14,219 training samples. The comparison therefore reflects a practical compute-constrained setting rather than a perfectly controlled architectural test.

Macro F1 tells a different story. BiLSTM reaches only 0.5478 because it collapses on the Neutral class, while Decision Tree maintains more balanced per-class performance. In practice, BiLSTM is preferable when overall Positive/Negative discrimination is the main goal, whereas ML+SMOTE is more reliable for balanced three-class classification.

\subsection{Discussion}

\subsubsection*{Effect of class imbalance}
Both tracks are affected by strong class imbalance. SMOTE helps the ML models retain Neutral-class performance, whereas the unweighted BiLSTM largely ignores that minority class.

\subsubsection*{Effect of training data asymmetry}
The 4.9$\times$ training-data asymmetry prevents a fully equal comparison. Even so, BiLSTM still shows a strong advantage on weighted metrics, suggesting that sequential context is useful for short Indonesian reviews.

\subsubsection*{Preprocessing effectiveness}
The shared preprocessing pipeline reduces vocabulary variation and keeps the final training vocabulary compact at 3,285 tokens. This benefits both TF-IDF features and embedding coverage.

\subsubsection*{Deployment and reproducibility}
Both the Decision Tree and BiLSTM models are deployed on Hugging Face Spaces, enabling direct inspection of their behavior.

\subsubsection*{Limitations}
The main limitations are training-data asymmetry, the absence of class balancing in the DL track, and different evaluation settings between ML cross-validation and DL hold-out testing. In addition, the study does not use pre-trained Indonesian embeddings.

% ==============================================================================
\section{Conclusion}
\label{sec:conclusion}

This study compares three classical machine learning classifiers with a BiLSTM for three-class sentiment classification of Indonesian Spotify reviews. With 5-fold cross-validation on the full cleaned dataset, Decision Tree is the strongest ML model with weighted F1-score 0.7269. On its held-out test set, BiLSTM reaches weighted F1-score 0.8069 and accuracy 0.8314.

The main trade-off lies in class balance. BiLSTM gives better overall weighted performance, but it fails on the minority Neutral class. In contrast, Decision Tree with SMOTE produces more stable performance across all three classes. Therefore, BiLSTM is preferable for overall polarity detection, while ML+SMOTE is more suitable when balanced three-class recognition is required.

Future work should train the BiLSTM on the full corpus, add class balancing to the DL pipeline, and test pre-trained Indonesian embeddings or transformer-based models.
% ==============================================================================

\bibliographystyle{unsrtnat} % Gaya sitasi unsrtnat umum untuk arXiv
\bibliography{references}    % Memanggil file references.bib

@article{pang2008opinion,
  title     = {Opinion Mining and Sentiment Analysis},
  author    = {Pang, Bo and Lee, Lillian},
  journal   = {Foundations and Trends in Information Retrieval},
  volume    = {2},
  number    = {1--2},
  pages     = {1--135},
  year      = {2008},
  doi       = {10.1561/1500000011}
}

@article{pratap2020comparative,
  title     = {Comparative Sentiment Analysis of App Reviews},
  author    = {Pratap, Ananya and Nidhi},
  journal   = {arXiv preprint arXiv:2006.09739},
  year      = {2020},
  url       = {https://arxiv.org/abs/2006.09739}
}

@article{rustam2021comparative,
  title     = {A Comparative Study of Sentiment Analysis Using {NLP} and Different
               Machine Learning Techniques on {US} Airline {Twitter} Data},
  author    = {Rustam, Furqan and Ashraf, Imran and Mehmood, Arif and
               Ullah, Saleem and others},
  journal   = {arXiv preprint arXiv:2110.00859},
  year      = {2021},
  url       = {https://arxiv.org/abs/2110.00859}
}

@article{hochreiter1997lstm,
  title     = {Long Short-Term Memory},
  author    = {Hochreiter, Sepp and Schmidhuber, J{\"u}rgen},
  journal   = {Neural Computation},
  volume    = {9},
  number    = {8},
  pages     = {1735--1780},
  year      = {1997},
  doi       = {10.1162/neco.1997.9.8.1735}
}

@article{xu2019bilstm,
  title     = {Sentiment Analysis of Comment Texts Based on {BiLSTM}},
  author    = {Xu, Guixian and Meng, Yuxin and Qiu, Xuxiang and
               Yu, Zhenlong and Wu, Xueliang},
  journal   = {IEEE Access},
  volume    = {7},
  pages     = {51522--51532},
  year      = {2019},
  doi       = {10.1109/ACCESS.2019.2911964}
}

@article{lin2023hybrid,
  title     = {Sentiment Analysis of {Indonesian} Datasets Based on a Hybrid
               Deep-Learning Strategy},
  author    = {Lin, Chih-Hsueh and Nuha, Ulin},
  journal   = {Journal of Big Data},
  volume    = {10},
  number    = {1},
  pages     = {88},
  year      = {2023},
  doi       = {10.1186/s40537-023-00782-9}
}

@inproceedings{wilie2020indonlu,
  title     = {{IndoNLU}: Benchmark and Resources for Evaluating {Indonesian}
               Natural Language Understanding},
  author    = {Wilie, Bryan and Vincentio, Karissa and Winata, Genta Indra and
               Cahyawijaya, Samuel and Li, Xiaohong and Lim, Zhi Yuan and
               Soleman, Sidik and Mahendra, Rahmad and Fung, Pascale and
               Bahar, Syafri and Purwarianti, Ayu},
  booktitle = {Proceedings of the 1st Conference of the Asia-Pacific Chapter of
               the Association for Computational Linguistics and the 10th
               International Joint Conference on Natural Language Processing},
  pages     = {843--857},
  year      = {2020},
  organization = {Association for Computational Linguistics},
  url       = {https://aclanthology.org/2020.aacl-main.85}
}

@article{kuncahyo2020bilstm,
  title     = {Improving {Bi-LSTM} Performance for {Indonesian} Sentiment Analysis
               Using Paragraph Vector},
  author    = {Kuncahyo, Setyo and others},
  journal   = {arXiv preprint arXiv:2009.05720},
  year      = {2020},
  url       = {https://arxiv.org/abs/2009.05720}
}

@article{farizki2025bilstm,
  title     = {Klasifikasi Sentimen Menggunakan Bidirectional {LSTM} dan {IndoBERT}
               dengan Dataset Terbatas},
  author    = {Farizki, Muhammad and others},
  journal   = {{ZONAsi}: Jurnal Sistem Informasi},
  year      = {2025},
  url       = {https://journal.unilak.ac.id/index.php/zn/article/view/25091}
}

@article{abiola2024sentiment,
  title     = {Sentiment Classification on the 2024 {Indonesian} Presidential
               Candidate Dataset Using Deep Learning Approaches},
  author    = {Abiola, Oluwaseun and others},
  journal   = {Indonesian Journal of Statistics and Its Applications},
  year      = {2024},
  url       = {https://journal-stats.ipb.ac.id/index.php/ijsa/article/view/1259}
}

@article{adriani2007stemming,
  title     = {Stemming {Indonesian}},
  author    = {Adriani, Mirna and Asian, Jelita and Nazief, Bobby and
               Tahaghoghi, Saied M. M. and Williams, Hugh E.},
  journal   = {{ACM} Transactions on Asian Language Information Processing},
  volume    = {6},
  number    = {4},
  pages     = {1--33},
  year      = {2007},
  doi       = {10.1145/1316457.1316458}
}

@article{ossai2024sentiment,
  title     = {Sentiment Analysis on {Google Play Store} App Users' Reviews
               Based on Deep Learning Approach},
  author    = {Ossai, Chukwuemeka Ivan and Wickramasinghe, Nilmini},
  journal   = {Multimedia Tools and Applications},
  year      = {2024},
  doi       = {10.1007/s11042-024-19185-w}
}

@article{chawla2002smote,
  title     = {{SMOTE}: Synthetic Minority Over-Sampling Technique},
  author    = {Chawla, Nitesh V. and Bowyer, Kevin W. and
               Hall, Lawrence O. and Kegelmeyer, W. Philip},
  journal   = {Journal of Artificial Intelligence Research},
  volume    = {16},
  pages     = {321--357},
  year      = {2002},
  doi       = {10.1613/jair.953}
}

\end{document}